\documentclass[10pt,twocolumn]{article}

\usepackage[utf8]{inputenc}
\usepackage[T1]{fontenc}
\usepackage[protrusion=true,expansion=false]{microtype}
\usepackage{amsmath,amssymb}
\usepackage{booktabs}
\usepackage{array}
\usepackage{url}
\usepackage[hidelinks]{hyperref}
\usepackage[margin=1in]{geometry}
\usepackage{titlesec}
\usepackage{authblk}

\titlespacing*{\section}{0pt}{1.2ex plus .2ex}{0.8ex}
\titlespacing*{\subsection}{0pt}{1.0ex plus .2ex}{0.6ex}
\newcommand{\prop}[1]{\textbf{(#1)}}

\title{\bfseries Reduced-Order Models: The Mother of World Models\\[0.4em]
\large Verification, Physical Grounding, and the Missing Half of the World-Model Agenda}

\author{Rajat Ghosh}
\affil{Independent Researcher \\ \texttt{rajat.ghosh11@gmail.com}}

\begin{document}
\twocolumn[
\maketitle
\begin{@twocolumnfalse}
\begin{abstract}
\noindent
World models --- compressed latent representations of an environment that
support action-conditioned prediction and planning --- are typically presented
as a product of modern self-supervised learning. This paper argues that the
functional anatomy of a world model was independently developed, deployed, and
formally analyzed decades earlier in the model-order-reduction (MOR) and control
literature, under different names and for a different purpose: the real-time
operation of physical systems. We trace the anatomy across three communities.
Low-dimensional models of turbulence built on proper orthogonal decomposition
(POD) supplied latent dynamics learned from data of a chaotic environment;
eigenface methods in early computer vision supplied the encoder--decoder half,
including a primitive runtime validity check; and measurement-based POD
frameworks for facility thermal control assembled the complete loop --- POD
coefficients as latent state, parametric dependence on actuator setpoints as
action conditioning, modal reconstruction as decoding, and --- critically ---
a priori analytical error bounds as a verification layer that certified when the
model's predictions could be trusted in closed loop. We then examine what each
tradition possesses that the other lacks: MOR contributes verification, physical
grounding, and extreme data efficiency; learned world models contribute
nonlinear representation, transferability, and horizon. We argue that the
outstanding obstacle to deploying world models in systems that cannot fail ---
power, thermal, process control --- is not predictive fidelity but verifiability,
and we outline a research agenda for physics-grounded, verifiable world models
that unifies the two lineages.
\end{abstract}
\vspace{1.5em}
\end{@twocolumnfalse}
]

\section{Introduction: Two Communities, One Architecture}

The autonomous operation of a physical system requires a model that can do three
coupled things: infer the system's latent state from partial observation, predict
how that state will evolve, and predict how it will evolve differently under
candidate interventions. Any agent --- human or artificial --- that operates a
data center, a distribution feeder, or a chemical process is running some version
of this loop, whether the model inside it is a senior operator's intuition, a
spreadsheet of rules of thumb, or a learned dynamics model queried by a planner.

Over the past eight years, one research community has given this architecture a
name and a research program. Beginning with the recurrent latent dynamics models
of Ha and Schmidhuber~\cite{ha2018}, continuing through the Dreamer line of
latent-imagination agents~\cite{dreamer1,dreamer2,dreamer3}, and generalized in
the joint-embedding predictive architectures advocated by LeCun~\cite{lecun2022},
the world model has become one of the organizing concepts of modern machine
learning: learn a compressed representation of the environment, learn its
dynamics in that representation, condition the dynamics on actions, and plan
against the model rather than the world. The term has become correspondingly
overloaded; a recent functional taxonomy from within the generative program
partitions the systems now called world models by their output --- renderers
emitting observations, simulators emitting state, planners emitting actions ---
and identifies the simulator, the physically faithful state-bearing tier, as the
least developed and most consequential of the three~\cite{li2026}. The program's
successes --- in games, simulated robotics, and increasingly in video-scale
generative models~\cite{genwm1,genwm2} --- have been driven by self-supervised
learning on large observational corpora, and its vocabulary (encoder, latent
state, rollout, imagination) reflects those origins.

This paper is about a second community that built the same architecture earlier,
for different reasons, in a different mathematical idiom --- and about what each
community has that the other still lacks. From the late 1990s through the 2010s,
the model-order-reduction (MOR) and control literature confronted a problem that
will sound familiar when stated in modern terms: high-fidelity models of physical
systems (computational fluid dynamics, finite-element analysis) were physically
grounded and rigorously convergent but orders of magnitude too slow for real-time
use, while the fast alternatives available to operators were heuristic and
unreliable. The response was to construct reduced representations ---
low-dimensional subspaces or manifolds capturing the dominant dynamics --- in
which prediction became fast enough for the control loop. Proper orthogonal
decomposition~\cite{lumley1967,sirovich1987}, balanced truncation~\cite{moore1981},
and reduced-basis methods~\cite{rozza2008,quarteroni2016} gave the community
principled encoders; Galerkin projection and interpolation gave it latent
dynamics; and --- the tradition's most distinctive achievement ---
functional-analytic error estimation gave it something the learned world-model
program still does not have: a priori bounds certifying, before a prediction is
consumed, how wrong it can be~\cite{rozza2008,quarteroni2016,benner2015}.

The two communities do not cite each other. A reader of the world-models
literature will search its bibliographies in vain for Sirovich, for reduced-basis
error estimators, or for the surrogate-assisted control literature; a reader of
the MOR literature will find no JEPA, no Dreamer, and no engagement with
self-supervised representation learning. This mutual invisibility is partly
sociological --- the communities publish in disjoint venues and optimize for
different benchmarks --- but it has a substantive cost in both directions. The
world-model program is converging, via safety concerns, on questions of
uncertainty, physical consistency, and certified deployment that the MOR
tradition spent two decades formalizing. The MOR tradition, meanwhile, hit hard
ceilings --- linearity of the representation, per-installation calibration, short
certified horizons --- that are precisely the ceilings modern representation
learning exists to break.

This paper makes three claims.

\smallskip
\noindent\textbf{Claim 1 (structural correspondence).} The anatomy of a world
model --- encoder, latent state, action-conditioned latent dynamics, decoder,
planner --- is isomorphic to the anatomy of a projection-based reduced-order
model deployed for control. This is not analogy; it is a component-by-component
identification, which we exhibit in \S3 across three instances from the
literature --- the low-dimensional turbulence models of the Lumley--Holmes
program, eigenface recognition in early computer vision, and a measurement-based
POD framework for facility thermal control that ran the complete
sense--predict--control loop on physical hardware, gated by an analytical error
bound.

\smallskip
\noindent\textbf{Claim 2 (the verification gap).} The property that determines
whether a predictive model may be placed inside the control loop of a system that
cannot fail is not average accuracy but verifiability --- the model's ability to
certify or bound its own predictions before they are acted upon. The MOR
tradition made verifiability its central artifact; the learned world-model
tradition has made comparatively little progress toward it, and this --- not
fidelity --- is the binding constraint on deployment in mission-critical domains
(\S4).

\smallskip
\noindent\textbf{Claim 3 (the synthesis agenda).} The limitations that retired
the classical reduced-order program --- linear subspaces defeated by
Kolmogorov-width barriers, empirically calibrated constants that do not transfer
across installations, certified horizons measured in seconds --- are exactly the
capabilities that self-supervised learning has since demonstrated (\S5).
Conversely, the guarantees the classical program possessed are exactly what
learned world models need to enter mission-critical loops. The intersection
defines a concrete research agenda: physics-grounded, verifiable world models
(\S6).

\smallskip
A remark on scope, to prevent a misreading. This paper makes no priority claim.
The modern world-model program was not derived from model-order reduction; the
lineages are independent, and the results of each stand on their own. Our title's
genealogy is anatomical, not historical: the reduced-order model is the mother of
the world model in the sense that it carried the same body plan first --- and in
the sense that what it knew about raising such models safely has not yet been
inherited.

\section{Mission-Critical World Models: An Operational Definition}

\subsection{The class of systems}

We use the term \emph{mission-critical physical system} for engineered systems in
which incorrect control action carries consequences that are severe, immediate,
and physical: thermal runaway in a compute facility, frequency instability in a
power network, off-specification excursion in a continuous chemical process, loss
of containment in an energy-storage installation. Three characteristics jointly
distinguish this class from the environments in which world models are typically
studied.

First, the cost function is asymmetric and discontinuous. In game-playing,
navigation, or video prediction, a poor prediction degrades reward smoothly; in a
mission-critical system, a single action that violates a physical constraint can
be unrecoverable. The relevant risk measure is not expected error but the
probability of any excursion beyond a hard envelope, and operators correspondingly
price model failure at close to infinity. This asymmetry explains an empirical
fact that has puzzled proponents of learned control: facilities with abundant
telemetry and clear optimization incentives continue to be operated by
conservative heuristics. The heuristics are not good; they are bounded, and
boundedness is what the deployment decision actually turns on.

Second, the data regime is inverted relative to the settings that produced modern
world models. Internet-scale corpora exist for text, images, and video; no such
corpus exists for the transient thermal behavior of a particular data hall or the
dynamic response of a specific distribution feeder, and none will. Each facility
is a distinct physical realization --- its geometry, equipment, degradation state,
and operating envelope are its own --- and the most informative regimes (faults,
limit excursions, cascading interactions) are precisely those a prudent operator
refuses to sample. A model class whose success depends on dense coverage of the
state--action space is structurally mismatched to environments where the
interesting states are the forbidden ones.

Third, the system is instrumented but partially observed, and it is actuated.
Sensors are sparse relative to the field variables that matter --- a few hundred
thermocouples against a continuous temperature field, a few thousand phasor
measurements against a continental network --- while the operator continuously
intervenes through setpoints, dispatch, and switching. Any model intended for
operation must therefore solve three coupled problems --- inferring the latent
physical state from sparse observation, predicting its evolution, and predicting
it conditioned on candidate interventions --- rather than the passive
next-frame-prediction problem that dominates the world-model literature.

\subsection{Four properties}

We define a mission-critical world model operationally, by what it must do rather
than by how it is built. The definition is deliberately
implementation-agnostic: nothing in it requires a neural network, and nothing in
it excludes one. A model of a physical system qualifies if and only if it
exhibits four properties simultaneously.

\prop{P1} \textbf{Physically grounded.} The model's predictions respect the
conservation structure and constitutive behavior of the target system. Grounding
may be achieved by construction (a basis derived from the governing equations, a
projection that preserves the energy balance), by constraint (physics-informed
losses or projections applied to a learned model~\cite{raissi2019,li2021fno}), or
by hybrid architecture; the property concerns the output, not the mechanism. A
model that can be induced to emit trajectories violating energy or mass balance
is not grounded, however photorealistic its rollouts. We emphasize that visual
plausibility and physical consistency are independent axes: contemporary
generative world models~\cite{genwm1,genwm2} occupy the high-plausibility,
low-consistency quadrant, which is acceptable for content creation and
disqualifying for control --- a limitation acknowledged from within the
generative program itself, whose own taxonomy concedes that renderers optimize
visual plausibility over physical accuracy and cannot be trusted to design a
building or train a robot~\cite{li2026}.

\prop{P2} \textbf{Verifiable.} The model can bound, certify, or at minimum flag
the validity of its own predictions before those predictions are consumed by a
controller. Verifiability admits degrees. The strongest form is an a priori
analytical bound --- a guarantee, derived from the structure of the
approximation, on the error of a prediction not yet made~\cite{rozza2008,quarteroni2016,benner2015}.
Weaker but still operationally meaningful forms include a posteriori residual
checks against the governing equations, conformal or calibrated uncertainty with
coverage guarantees~\cite{vovk2005,angelopoulos2023}, and runtime-assurance
envelopes that gate a learned policy behind a certified
monitor~\cite{sha2001,ames2019}. What does not qualify is uncalibrated
confidence: an ensemble variance or a softmax temperature is a diagnostic, not a
certificate. We will argue (\S4) that verifiability is the property on which
deployment actually turns, and the one the learned world-model program has made
the least progress toward.

\prop{P3} \textbf{Real-time.} Inference must fit inside the control loop of the
target system --- milliseconds to seconds for power-electronic and rack-level
thermal dynamics, seconds to minutes for facility-level thermal and process
dynamics. This requirement excludes the classical alternative to learned models:
first-principles numerical simulation achieves grounding and, through convergence
theory, a form of verifiability, but at wall-clock costs of hours to days per
scenario~\cite{simcost,dtwin}. The entire motivation for reduced and learned
representations in this setting is to preserve as much of P1 and P2 as possible
while satisfying P3.

\prop{P4} \textbf{Causal and controllable.} The model predicts the consequences
of interventions, not merely the continuation of observed trajectories: its
inputs include the actuation variables, and its predictions respond to them with
the correct sign, magnitude, and transient structure. This is the property that
separates a world model from a forecaster. A model trained purely on passive
observation of a system under closed-loop control learns the closed-loop
correlational structure --- including the confounding induced by the incumbent
controller --- and will generally mispredict the effect of actions outside the
historical policy~\cite{pearl2009}. Action-conditioning must therefore be
architectural and, ideally, validated by intervention.

Two remarks on the definition. First, the four properties are individually
attainable and jointly hard: numerical simulation achieves P1, P2, and P4 but
fails P3; generative video models achieve P3 and arguably a weak form of P4 but
fail P1 and P2; correlational surrogates and dashboard digital twins~\cite{dtwin}
achieve P3 but fail P2 and typically P4. The class of interest is the
intersection, and the central claim of this paper is that the intersection has
been reached at least once before (\S3), under assumptions whose relaxation
defines a research agenda (\S6).

Second, the definition induces an operational acceptance test that we will use
throughout:

\begin{quote}
A model qualifies for a mission-critical loop when, for every action it proposes,
it can either predict the consequence within a certified tolerance or declare
that it cannot. Silent failure --- confident prediction outside the model's
region of validity --- is the disqualifying behavior.
\end{quote}

The test is deliberately asymmetric: it does not demand accuracy everywhere, only
honesty everywhere. A model with a modest certified envelope and a reliable
refusal boundary is deployable; a model with superior average accuracy and no
knowledge of its own boundary is not. This inversion of the usual accuracy-first
evaluation ordering is, we contend, the correct ordering for physical systems
that cannot fail, and it is the ordering under which the reduced-order tradition
--- whose central artifacts were error bounds rather than benchmarks --- is most
instructive.

\subsection{What a mission-critical world model is not}

The definition excludes, deliberately, three neighboring constructions with which
it is often conflated. It is not a generative world model: photorealistic or
3D-consistent scene synthesis~\cite{genwm1,genwm2} optimizes perceptual
plausibility, which is neither necessary nor sufficient for P1--P2. It is not a
general latent planner: abstract world models for open-ended
reasoning~\cite{lecun2022} pursue task-agnostic representation, whereas the
mission-critical setting demands fidelity to one specific physical system,
verified. And it is not a digital twin in the prevailing industrial sense of a
synchronized monitoring replica~\cite{dtwin}: visualization and correlational
anomaly detection satisfy neither P2 nor P4. These exclusions are not criticisms
of the excluded programs --- each is well-matched to its own objective --- but
boundary markers: the requirements of systems that cannot fail define a distinct
point in the design space, and it is that point this paper addresses.

It is instructive to locate this definition against the functional taxonomy of
Li~\cite{li2026}, which partitions world models by what they output: renderers
(observations), simulators (state), planners (actions). The two schemes are
orthogonal. The taxonomy classifies by output modality; the definition of \S2.2
classifies by guarantee. In the taxonomy's terms, a mission-critical world model
is a simulator--planner composition --- it must carry physically faithful state
(the simulator's contract) and emit actions against it (the planner's contract).
But the comparison also exposes the gap this paper is about: none of the three
functional categories has a column for the model knowing when it is wrong.
Verification is absent from the taxonomy not by oversight but because it is absent
from the systems the taxonomy describes --- which is precisely Claim 2, stated by
omission.

\section{Case Studies: One Anatomy, Three Communities}

The correspondence asserted in Claim 1 is not the property of a single artifact;
it is a pattern that recurs wherever empirical low-dimensional representation met
the demands of a real environment. We exhibit three instances in roughly
chronological order. The first, from turbulence research, built the latent
dynamics; the second, from computer vision, built the encoder--decoder --- and,
less remembered, a primitive validity check; the third, from facility thermal
management, assembled every component of the anatomy, including the one the
modern program still lacks, in a single closed loop.

\subsection{Turbulence: latent dynamics of a chaotic environment}

The oldest instance is also the most direct ancestor of the latent-dynamics idea.
Lumley proposed proper orthogonal decomposition as a means of extracting the
coherent structures of turbulence --- the energetically dominant, recurring
patterns hiding in an apparently disordered field~\cite{lumley1967} --- and
Sirovich's method of snapshots made the decomposition computable from data
ensembles~\cite{sirovich1987}. The decisive step came when Aubry, Holmes, Lumley,
and Stone projected the Navier--Stokes equations onto a handful of empirical
modes of the turbulent boundary layer, obtaining a low-dimensional system of
ordinary differential equations whose trajectories reproduced qualitative regimes
--- intermittency, bursting --- of the wall region~\cite{aubry1988}; the program
was codified in Holmes, Lumley, and Berkooz~\cite{holmes1996}.

Restated in modern vocabulary, this is a world model of a chaotic physical
environment, minus actions: an encoder (projection onto modes learned from data),
latent dynamics (the Galerkin ODEs evolving the coefficients), and a decoder
(modal reconstruction of the flow field). The representation was learned from
observations of the environment; the dynamics were compact enough to analyze and
integrate in real time; and the object of the exercise --- predict the evolution
of a high-dimensional system in a low-dimensional latent space --- is precisely
the object of latent imagination. Action conditioning arrived when the
construction was turned toward flow control: POD-based reduced models conditioned
on actuation were used to design and optimize feedback control of wake
flows~\cite{noack2003,bergmann2005}, completing P4 for this lineage. What the
turbulence program never possessed was a deployed verification layer; indeed its
best-known failure mode --- empirical modes deforming under the very actuation the
model was asked to predict --- is an early, instructive instance of the
distribution-shift problem that motivates \S4.

\subsection{Vision: the encoder--decoder half, and a primitive verifier}

The same mathematical machinery crossed into perception. Sirovich and Kirby
showed that human faces occupy a low-dimensional subspace: an empirical orthogonal
basis learned from an ensemble of face images compresses a face to a coefficient
vector of modest dimension and reconstructs it recognizably~\cite{sirovich1987kirby}.
Turk and Pentland turned the representation into a recognition system ---
eigenfaces --- by classifying in the latent space~\cite{turk1991}. In modern
terms this is exactly the encoder--decoder half of a world model, learned from
data, in the domain (vision) whose later successes would drive deep
representation learning; the eigenface subspace is the direct intellectual
ancestor of the learned embedding.

One detail of the eigenface system deserves more attention than it receives. Turk
and Pentland used the reconstruction residual --- the distance between an input
image and its projection onto the face subspace --- as a runtime test of whether
the input was a face at all, rejecting inputs the representation could not
explain~\cite{turk1991}. This is a validity check of the model's own
applicability, consumed at inference time: a primitive, purely geometric ancestor
of property P2. The perception community built a self-verifying encoder in 1991
and then, in the deep-learning transition, largely lost the habit.

\subsection{The complete loop: facility thermal control}

The third instance assembled the full anatomy. Drawn from the reduced-order
thermal-management literature of the early 2010s, it is a measurement-based POD
framework for the transient thermal operation of data centers, validated in
closed loop in a research facility~\cite{ghosh_thermal,ghosh_error}, and it sits
within a broader body of reduced-order data center modeling from the same
period~\cite{rambo,samadiani,cfddc}. We single it out because it is unusually
complete: a single framework performing state estimation from sparse sensing,
action-conditioned prediction, closed-loop control on physical hardware, and a
priori error certification --- every component of the modern anatomy, in one
artifact.

The physical problem is representative of the mission-critical class of \S2.1. A
data center's air temperature field is a continuous, convection-dominated
distributed state, observed through a sparse grid of thermocouples (on the order
of one hundred measurement points in the validation facility) and actuated
through the setpoints of computer-room air-conditioning (CRAC) units against
time-varying heat loads. Overcooling wastes a large fraction of facility energy;
undercooling risks thermal excursions with hard consequences. The incumbent
practice --- conservative static setpoints with wide margins --- is precisely the
bounded heuristic of \S2.1. A design decision matters for the correspondence: the
reduced basis was constructed from measured temperature data rather than from
simulation, so the modes encoded the facility as it actually behaved --- leakage,
recirculation, equipment idiosyncrasy included. The representation was learned
from observational data of the target environment, not from a simulator of it.

Each component maps onto the anatomy of \S2.2; Table~\ref{tab:correspondence}
summarizes, and four identifications deserve elaboration.

\smallskip
\noindent\textit{Latent state and decoder.} The empirical eigenmodes of the
measured ensemble were strongly low-rank: a small number of modes captured the
overwhelming majority of the thermal field's energy, so that the state of a
facility-scale distributed system was carried by a coefficient vector of
dimension ten or less. The decoder --- linear reconstruction through the modes
--- did more than reproduce the sensed points: it estimated the field between
sensors, converting sparse observation into dense state. This is state estimation
in the observer-theoretic sense, and it is the capability that modern world-model
treatments of partially observed physical systems must reproduce with encoders.

\smallskip
\noindent\textit{Action conditioning.} The framework was parametric in the
actuation variables: the reduced dynamics were constructed as functions of CRAC
setpoint and heat load, so that querying the model with a candidate setpoint
yielded the predicted transient response to that intervention. The model was
therefore not a forecaster of the closed-loop trajectory but a conditional model
of intervention outcomes --- property P4 --- and this was validated
interventionally, by actuating the physical facility and comparing.

\smallskip
\noindent\textit{The verifier.} The framework's most distinctive component, and
the one with no counterpart in the modern program, was an a priori error estimate
derived from the functional-analytic structure of the approximation: before a
prediction was made, the framework produced a bound on its uncertainty as a
function of where the query lay relative to the training ensemble. Within the
interpolation regime --- queries parametrically interior to the observed ensemble
--- the certified uncertainty was on the order of a few percent at high
confidence. In extrapolation, the bound degraded gracefully and knowably:
uncertainty grew to roughly ten percent, with worst-case local deviations larger
still, and the framework defined an explicit temporal horizon (on the order of
tens of seconds at the chosen tolerance) beyond which predictions were not to be
trusted. Crucially, the bound was not a diagnostic reported after the fact; it
was consumed upstream of action.

\smallskip
\noindent\textit{The closed loop.} The full architecture was exercised end-to-end
on physical hardware: sparse sensing to latent state, action-conditioned
prediction over candidate setpoints, optimization in the reduced space, actuation
of the selected setpoint, and re-verification --- with demonstrated cooling-energy
reductions of tens of percent relative to the conservative baseline, at prediction
errors of a few percent~\cite{ghosh_thermal,ghosh_error}. By the operational test
of \S2.2, the system qualified: it acted where certified and declined where not.

\subsection{Honest accounting}

These case studies are offered as existence proofs, not solutions, and their
limitations must be stated with the same precision as their achievements --- they
are the subject of \S5 and the motivation for \S6. We account for the complete
instance, whose limitations subsume the others'.

The representation was linear: a fixed subspace spanned by empirical modes. Its
success depended on the low-rank character of the facility's dominant transients,
and linear subspaces are known to fail --- by Kolmogorov-width arguments --- on
transport-dominated and strongly nonlinear regimes~\cite{ohlberger2016}; the
turbulence lineage met exactly this wall, along with mode deformation under
actuation (\S3.1). The framework's calibration was installation-specific:
empirical constants in the reduced dynamics were identified from benchmarking
experiments in the target facility, and nothing in the construction transferred
to a different facility without repeating that identification. The certified
horizon was short: tens of seconds of trustworthy extrapolation is sufficient for
setpoint regulation and short-term prognostics, and insufficient for the
longer-horizon planning that motivates the modern program. And the validation,
while interventional and hardware-in-the-loop, was confined to a single facility
and a single air-delivery scheme; generality across facility classes was asserted
by argument, not demonstrated by replication.

One of these limitations deserves a dual reading. The short certified horizon is
a limitation of scope but a triumph of epistemics: the model did not fail silently
at the boundary of its validity --- it declared that boundary in advance,
analytically. No learned world model known to us can currently make the
corresponding declaration. The classical program thus sits in a curious position
relative to the modern one: strictly weaker in representational power, and
strictly stronger in self-knowledge.

\begin{table*}[t]
\centering
\small
\begin{tabular}{@{}p{0.28\textwidth}p{0.64\textwidth}@{}}
\toprule
\textbf{World-model component} & \textbf{Measurement-based POD framework} \\
\midrule
Encoder $\rightarrow$ latent state & Projection of the measured field onto POD
modes; $\mathcal{O}(10)$ coefficients summarize $\mathcal{O}(100)$ sensor
readings \\
Latent dynamics & Temporal evolution of the POD coefficients \\
Action conditioning & Parametric dependence of the coefficient dynamics on CRAC
setpoint and heat load \\
Decoder & Modal reconstruction of the full temperature field, including at
unsensed locations \\
Planner / controller & Iterative setpoint optimization performed in the reduced
space; closed loop on hardware \\
Verifier & A priori analytical error bound; certified interpolation regime and a
defined extrapolation horizon \\
\bottomrule
\end{tabular}
\caption{Component-by-component correspondence between the modern world-model
anatomy and the measurement-based reduced-order framework for facility thermal
control of \S3.3~\cite{ghosh_thermal,ghosh_error}.}
\label{tab:correspondence}
\end{table*}

\begin{table}[t]
\centering
\small
\begin{tabular}{@{}lccc@{}}
\toprule
\textbf{Component} & \textbf{Turb.} & \textbf{Faces} & \textbf{Thermal} \\
\midrule
Encoder / latent state & $\bullet$ & $\bullet$ & $\bullet$ \\
Latent dynamics        & $\bullet$ & ---       & $\bullet$ \\
Decoder                & $\bullet$ & $\bullet$ & $\bullet$ \\
Action conditioning    & $\circ$   & ---       & $\bullet$ \\
Closed-loop control    & $\circ$   & ---       & $\bullet$ \\
Verification           & ---       & $\circ$   & $\bullet$ \\
\bottomrule
\end{tabular}
\caption{The anatomy assembled across the three case studies ($\bullet$ present;
$\circ$ partial --- action conditioning and control in the turbulence lineage
arrived with the flow-control extensions~\cite{noack2003,bergmann2005};
verification in the eigenface system was the geometric residual test of \S3.2).}
\label{tab:anatomy}
\end{table}

\section{What the Reduced-Order Tradition Has That Learned World Models Need}

\subsection{Verification as the deployment-binding constraint}

The learned world-model literature evaluates on predictive fidelity: rollout
accuracy, downstream task reward, perceptual quality. These orderings are
appropriate to their settings. But the deployment decision for a mission-critical
system is not made on fidelity --- it is made on the question \emph{what happens
when the model is wrong, and will we know beforehand?} An operator who accepts a
model into the loop is underwriting its worst case, not its average case, and
rationally demands a certificate rather than a scoreboard.

The reduced-order tradition organized itself around exactly this demand. Its
signature artifacts are not benchmarks but bounds: a priori error estimates for
reduced-basis approximations~\cite{rozza2008,quarteroni2016}, computable error
bounds for balanced truncation~\cite{moore1981}, residual-based a posteriori
estimators that convert the governing equations into runtime
checks~\cite{benner2015}. The intellectual habit --- every approximation ships
with an estimate of its own error --- was not an optional refinement; it was the
community's admission ticket to engineering practice.

The learned program's nearest analogues are materially weaker. Deep ensembles and
Bayesian approximations quantify epistemic spread, which correlates with error but
certifies nothing. Conformal methods~\cite{vovk2005,angelopoulos2023} provide
genuine coverage guarantees and represent the most promising bridge, but their
guarantees are distributional --- contingent on exchangeability with the
calibration data --- and therefore weakest precisely where mission-critical
systems need them most: under the distribution shift induced by novel
interventions and excursions. Runtime assurance architectures~\cite{sha2001,ames2019}
gate learned policies behind certified monitors, which is the correct
systems-level pattern, but they presuppose a verifiable monitor --- which returns
the problem to where the ROM tradition left it. The most rigorous guarantee yet
produced by the learned program is instructive about the distance remaining. A
recent identifiability theorem for the joint-embedding line proves that a LeJEPA
encoder linearly recovers the world's true latent variables --- provided the
latents are Gaussian and evolve under stationary, additive-noise
transitions~\cite{klindt2026,balestriero2025}. Two observations follow. First,
this is a representation guarantee --- structural correctness of the learned
latent, established at training time --- not a runtime, query-conditioned
certificate of predictive validity; it constrains what the model is, not how
wrong any particular prediction may be. Second, the assumptions under which the
guarantee holds --- Gaussian latents with stationary additive-noise dynamics ---
describe a world nearly as restrictive as the linear regimes in which classical
reduced-order guarantees lived. The learned program's first formal guarantee, in
other words, was purchased at the classical program's prices. The gap is real:
there is at present no accepted construction that gives a neural dynamics model
the a priori, query-conditioned validity certificate that a classical linear ROM
carried as standard equipment.

\subsection{Physical grounding by construction}

A projection-based ROM inherits structure from the equations it reduces:
conservation properties, stability characteristics, and symmetries can be
preserved by the projection or enforced in the reduced system~\cite{structpres}.
The model cannot be prompted, perturbed, or extrapolated into emitting a
trajectory that violates the energy balance, because the balance is load-bearing
in the construction. Learned world models enforce physical consistency, when they
enforce it at all, through soft penalties~\cite{raissi2019} or architectural
priors~\cite{li2021fno}, and their failure mode under distribution shift is
characteristic: fluent, plausible, physically impossible rollouts. For content
generation this is a quality problem; inside a control loop it is the silent
failure that the test of \S2.2 disqualifies. The ROM lesson is not that learned
models must be projections --- it is that grounding enforced by construction is
categorically more trustworthy than grounding encouraged by loss, and that hybrid
constructions should preserve as much load-bearing structure as representational
ambition allows.

\subsection{Data efficiency where corpora do not exist}

The thermal-control framework of \S3.3 was trained on ensembles of measured
fields from one facility --- by modern standards, a vanishingly small dataset ---
and achieved certified few-percent accuracy, because the physics guaranteed that
the relevant dynamics occupied a low-dimensional manifold and the representation
was built to exploit exactly that. This is the general MOR bargain: accept a
strong prior (the governing structure), and the data requirement collapses by
orders of magnitude. In the mission-critical setting this bargain is not merely
economical but existential: as argued in \S2.1, the internet-scale corpus does
not exist for a specific facility and cannot be collected, because the informative
regimes are the forbidden ones. Physics is not a regularizer of convenience here;
it is the substitute for the data one is not permitted to gather. Any world-model
program aimed at these systems that does not make its physics prior do
quantitative, load-bearing work is implicitly assuming a data regime that will
never arrive.

\section{What Learning Has That the Reduced-Order Tradition Could Not Achieve}

The previous section is one half of a symmetric ledger. The classical program did
not stall for lack of rigor; it stalled against three ceilings that are, in
retrospect, precisely the capabilities that self-supervised representation
learning has since demonstrated. Intellectual honesty about these ceilings is
what separates the synthesis of \S6 from nostalgia.

\subsection{Nonlinearity and the width barrier}

The POD/reduced-basis construction approximates the solution set with a linear
subspace, and its efficiency is governed by the Kolmogorov $n$-width of that set:
when the width decays slowly --- as it provably does for transport-dominated,
advection-driven, and strongly nonlinear phenomena --- no linear subspace of
tractable dimension suffices~\cite{ohlberger2016}. The thermal instance of \S3.3
succeeded because a data hall's dominant thermal transients happened to be
low-rank; the same construction aimed at a strongly convective regime, a
multiphase process, or a network in a cascading contingency would meet the width
barrier head-on. The classical community's responses --- local bases, nonlinear
registration, manifold interpolation --- were partial. Learned nonlinear
representations dissolve the barrier in principle: an autoencoder or
joint-embedding architecture is not confined to a subspace, and the empirical
success of neural operators on parametric PDE families~\cite{li2021fno,lu2021}
demonstrates nonlinear solution manifolds captured at accuracy--speed points no
linear method reaches. Representation is the argument the learned program wins
outright.

\subsection{Transfer and the calibration ceiling}

The classical framework's empirical constants were identified per installation, by
benchmarking experiments in the target facility, and this was not an
implementation shortcut but a structural feature of the approach: the reduced
model was a bespoke artifact of one physical realization. The cost of standing up
each new deployment therefore scaled linearly with the number of deployments ---
an acceptable economics for one research facility and a fatal one for a fleet. The
modern program's central economic discovery is amortization: pretrain on broad
related data, adapt cheaply to the instance. A learned dynamics model pretrained
across many facilities' telemetry and many simulated geometries, then calibrated
to a target installation from days rather than months of data, inverts the
classical cost structure. Calibration itself becomes a learning problem --- and
the classical per-site constants are reinterpreted as the adaptation parameters of
a foundation model. Nothing in the classical toolkit offered this; it is the
second argument learning wins.

\subsection{Horizon}

The classical framework's certified extrapolation horizon was tens of seconds ---
enough for regulation and short-term prognostics, and an order of magnitude short
of the planning horizons that make a world model strategically interesting
(contingency evaluation, scenario rollout, what-if analysis over operational
timescales). Latent-space rollout with learned dynamics has demonstrated stable
long-horizon imagination in domains far harder to model
linearly~\cite{dreamer1,dreamer2,dreamer3}, through mechanisms --- stochastic
latent transitions, self-supervised regularization of the latent geometry,
rollout-consistency objectives --- that have no classical counterpart. The honest
caveat from \S4 must be attached: these long horizons are empirical, not
certified, and a long uncertified horizon is not yet an operational asset in the
mission-critical setting. But the representational capacity for long-horizon
prediction exists on one side of the ledger and not the other, and closing the
certification gap for horizons the classical program could never reach at all is a
better problem than possessing certificates for horizons too short to matter.

\section{A Research Agenda: Physics-Grounded, Verifiable World Models}

The ledger of \S\S4--5 is complementary by inspection: each tradition's strengths
sit opposite the other's ceilings. We state the synthesis as five open problems.
Each is posed as a research question with a concrete success criterion; none, to
our knowledge, is solved.

\smallskip
\noindent\textbf{Problem 1 --- Hybrid representation: learned dynamics on physical
structure.} Construct latent dynamics models whose representational class is
nonlinear (escaping the width barrier of \S5.1) but whose predictions preserve
load-bearing physical structure by construction rather than by penalty (retaining
\S4.2). Candidate directions include structure-preserving learned projections,
conservation-constrained latent transitions, and architectures in which a learned
nonlinear encoder feeds dynamics constrained to a physically interpretable
manifold. \emph{Success criterion:} a model that demonstrably cannot emit
balance-violating rollouts, at accuracy exceeding any linear ROM of comparable
inference cost, on a transport-dominated benchmark where POD provably fails.

\smallskip
\noindent\textbf{Problem 2 --- Verification for learned dynamics.} Extend the a
priori and a posteriori machinery of the reduced-order tradition to neural
dynamics models: query-conditioned validity certificates that hold under the
distribution shift induced by novel interventions. Plausible ingredients include
physics-residual monitors (the governing equations as a runtime oracle),
operator-theoretic perturbation bounds for restricted architecture classes,
conformal methods made shift-aware through physics-informed nonconformity scores,
and runtime-assurance envelopes whose certified monitor is itself a verified
reduced model --- a construction in which the classical ROM is not replaced by the
learned model but becomes its guardian. \emph{Success criterion:} a certificate
that, on held-out interventions outside the training policy, bounds realized error
with stated coverage --- and whose refusals are informative rather than vacuous.

\smallskip
\noindent\textbf{Problem 3 --- Calibration as transfer learning.} Reinterpret the
classical per-installation calibration constants (\S5.2) as the adaptation
parameters of a pretrained model, and characterize the sample complexity of
instance adaptation: how many days of telemetry, and which excitations, suffice to
certify a pretrained dynamics model on a new facility? The question has a
classical flavor --- it is optimal experiment design married to transfer learning
--- and a sharp economic consequence, since it governs whether physics-grounded
world models deploy as bespoke artifacts or as a platform. \emph{Success
criterion:} demonstrated cross-facility transfer in which adaptation data
requirements are reduced by an order of magnitude relative to from-scratch
identification, with certification (Problem 2) surviving the transfer.

\smallskip
\noindent\textbf{Problem 4 --- The simulation--measurement bargain.} The
thermal-control framework of \S3.3 chose measurement over simulation for
grounding; the data economics of Problem 3 push toward simulation for volume. The
correct architecture is plausibly both --- cheap simulated ensembles for
pretraining breadth, sparse measured data for grounding and validation --- but the
theory of this mixture is absent: when does simulation pretraining help versus
imprint simulator bias, what measured excitation is sufficient to correct it, and
how should a verifier treat predictions supported only by simulated evidence?
\emph{Success criterion:} a characterization, empirical or theoretical, of the
sim-to-real error budget for at least one mission-critical system class, with the
verifier of Problem 2 distinguishing simulation-supported from
measurement-supported certainty.

\smallskip
\noindent\textbf{Problem 5 --- Benchmarks that measure honesty.} The field
evaluates world models by fidelity; \S2.2 argued that mission-critical deployment
turns on honesty. A benchmark suite for this setting must therefore score,
alongside rollout accuracy: calibration of self-declared validity, refusal
behavior at the boundary of competence, action-conditioned (interventional)
accuracy rather than passive forecasting, and physical-consistency violation rates
under adversarial querying. Recent stress-testing gives the motivation empirical
teeth: systematic perturbation studies report that leading world-model
architectures degrade sharply under minor visual shifts as small as an agent color
change~\cite{stableworldmodel} --- evidence that fidelity-ranked leaderboards
overstate operational readiness. Candidate substrates --- facility thermal
dynamics, distribution-feeder response, continuous-process units --- are well
instrumented and physically well understood, making ground truth tractable.
\emph{Success criterion:} a public benchmark on which a verifiably honest model
with modest accuracy can legitimately outrank a more accurate model that fails
silently --- because that ordering, we have argued, is the ordering that governs
reality.

We note a systems-level implication of Problem 2 worth making explicit. The most
achievable near-term synthesis may not be a single monolithic model but a two-tier
architecture: a high-capacity learned world model proposing, and a classically
verifiable reduced model disposing --- certifying, bounding, or vetoing each
proposed action within its own certified envelope. In such a construction the
reduced-order model is not the ancestor the field outgrew; it is the component
that lets its descendant be trusted.

\section{Conclusion}

The name is new; the architecture is not. A decade before the term \emph{world
model} organized a research program, the model-order-reduction and control
community built compressed latent representations of physical systems, evolved
them forward conditioned on actions, decoded them to full fields, closed control
loops against them on real hardware --- and, distinctively, shipped every
prediction with a certificate of its own validity. That tradition ran into
ceilings --- linear representation, per-installation calibration, short certified
horizons --- that modern self-supervised learning has since shown how to break;
the modern program, in turn, lacks precisely the property, verification, on which
admission to mission-critical loops depends. Neither community closes the gap
alone. The systems that most need world models --- the power, thermal, and
process infrastructure on which everything else runs --- will be operated by
models that inherit from both: the representational reach of the learned
tradition, and the epistemic conscience of the reduced-order one.



\begin{thebibliography}{99}
\bibitem{genwm1} [Representative video-scale generative world-model citation ---
to be finalized.]
\bibitem{genwm2} [Representative 3D-consistent world-generation citation --- to be
finalized.]
\bibitem{structpres} [Representative structure-preserving model-reduction citation
--- to be finalized.]
\bibitem{samadiani} E. Samadiani and Y. Joshi. Proper orthogonal decomposition for
reduced-order thermal modeling of air-cooled data centers. [Venue/year to be
verified.]
\bibitem{cfddc} [Representative CFD-based data center thermal modeling citation ---
to be finalized.]
\bibitem{ghosh_thermal} R. Ghosh and Y. Joshi. [Measurement-based POD framework
for rapid transient temperature prediction in data centers --- exact citation to
be verified.]
\bibitem{ghosh_error} R. Ghosh and Y. Joshi. [Error estimation and uncertainty
bounds for POD-based reduced-order thermal prediction --- exact citation to be
verified.]
\end{thebibliography}
\end{document}